\documentclass[10pt]{article}

\usepackage[cmex10]{amsmath}
\usepackage{amsthm}
\usepackage{amsfonts}
\usepackage{amssymb}
\usepackage{url}
\usepackage{graphicx}
\usepackage{subfigure}
\usepackage{caption}
\newcommand{\RN}[1]{\textup{\uppercase\expandafter{\romannumeral#1}}}
\graphicspath{{./figure/}}

\begin{document}

\title{Challenges of Feature Selection for Big Data Analytics\footnote{It is a preprint version. The final version is to appear in Special Issue on Big Data, IEEE Intelligent Systems.}}
\author{
  Jundong Li and Huan Liu\\
  Computer Science and Engineering\\
  Arizona State University, USA\\
  \texttt{\{jundongl,huan.liu\}@asu.edu}
}
\date{}
\maketitle

\begin{abstract}

We are surrounded by huge amounts of large-scale high dimensional data. It is desirable to reduce the dimensionality of data for many learning tasks due to the curse of dimensionality. Feature selection has shown its effectiveness in many applications by building simpler and more comprehensive model, improving learning performance, and preparing clean, understandable data. Recently, some unique characteristics of big data such as data velocity and data variety present challenges to the feature selection problem. In this paper, we envision these challenges of feature selection for big data analytics. In particular, we first give a brief introduction about feature selection and then detail the challenges of feature selection for structured, heterogeneous and streaming data as well as its scalability and stability issues. At last, to facilitate and promote the feature selection research, we present an open-source feature selection repository (scikit-feature), which consists of most of current popular feature selection algorithms.

\end{abstract}

\noindent \textbf{Keywords.} Feature Selection; Big Data; Repository

%

\section{A Brief Introduction of Feature Selection}
\begin{figure}[!t]
\centering
\begin{minipage}{0.75\textwidth}
\centering
\subfigure[feature number growth]
{\includegraphics[width=\textwidth]{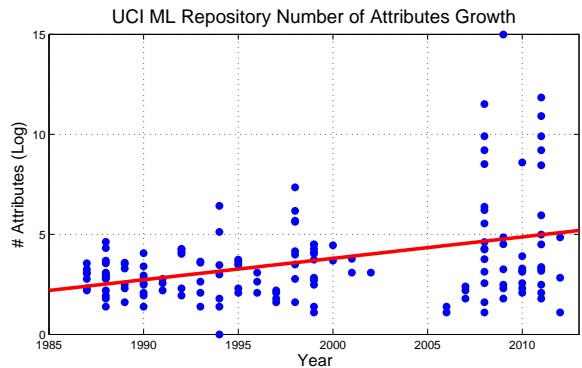}}
\end{minipage}
\begin{minipage}{0.75\textwidth}
\centering
\subfigure[sample size growth]
{\includegraphics[width=\textwidth]{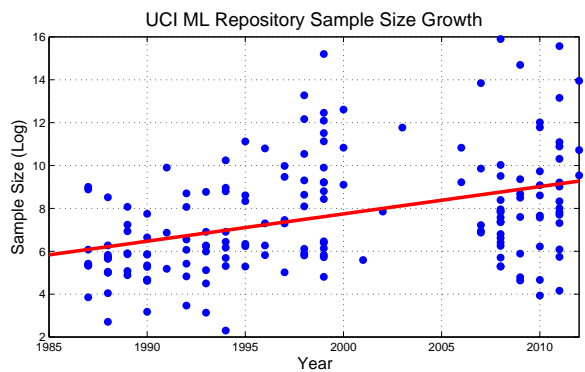}}
\end{minipage}
\centering
\caption{Samples and features growth trend during the past thirty years in the UCI machine learning repository.}
\label{fig:UCI}
\end{figure}

Massive amounts of high dimensional data are pervasive in multiple different domains, ranging from social media, e-commerce, bioinformatics, health care, transportation to online education. As an example, we show the growth trend of instance numbers and feature numbers in the UCI machine learning repository~\cite{bache2013uci} in Figure~\ref{fig:UCI}. As can be observed, both the data sample size and feature numbers are continuously growing over time. When applying data mining and machine learning algorithms on high dimensional data, a critical issue is known as curse of dimensionality. It refers to the phenomenon that data becomes sparser in high dimensional space, adversely affecting algorithms designed for low dimensional space. In addition, the existence of high dimensional features will significantly demand more on the computational and memory storage requirements.

\begin{figure}[!t]
\centering
\begin{minipage}{0.4\textwidth}
\centering
\subfigure[relevant feature $f_{1}$\label{fig:featureIllustration-a}]
{\includegraphics[width=\textwidth]{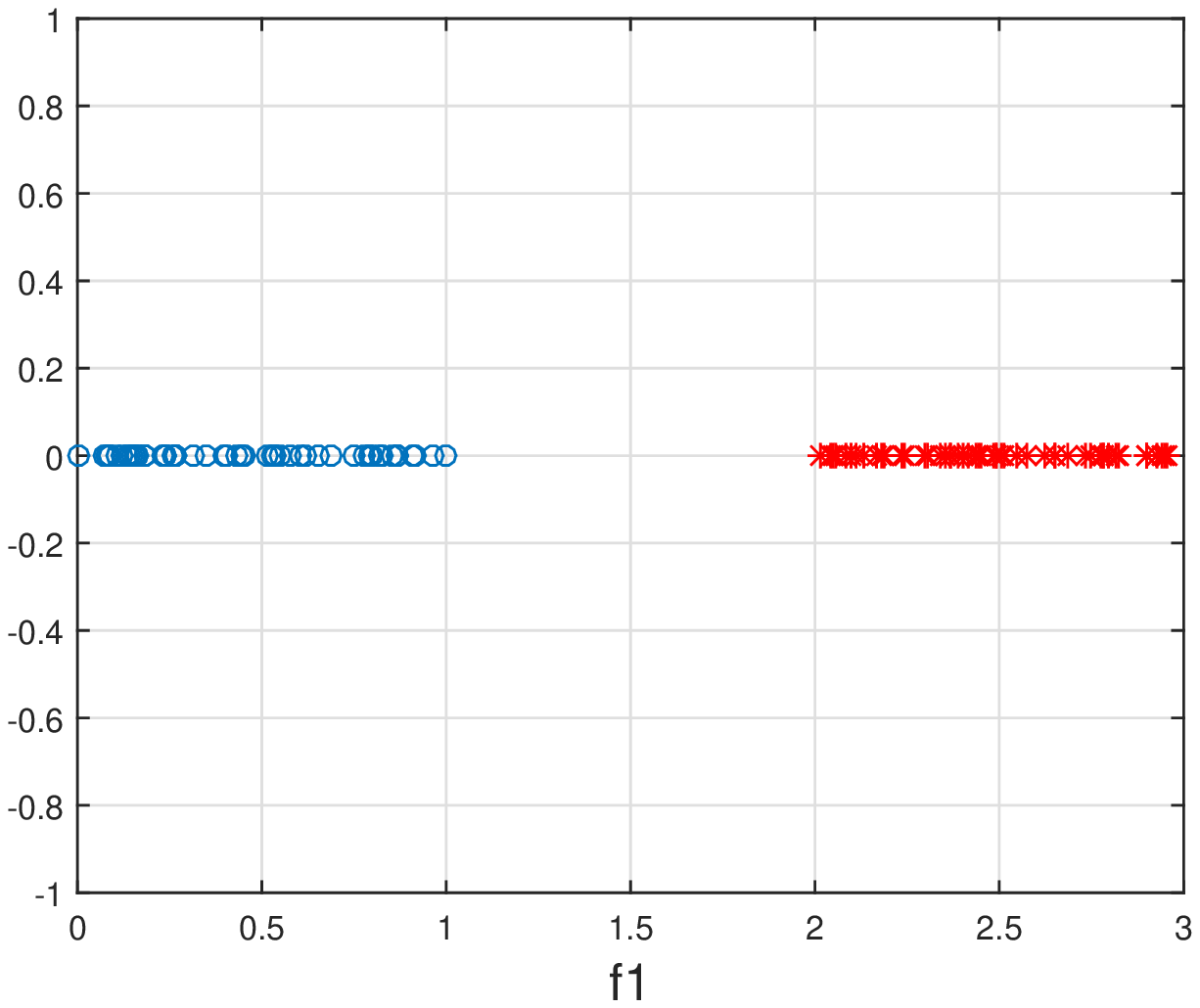}}
\end{minipage}
\begin{minipage}{0.4\textwidth}
\centering
\subfigure[redundant feature $f_{2}$\label{fig:featureIllustration-b}]
{\includegraphics[width=\textwidth]{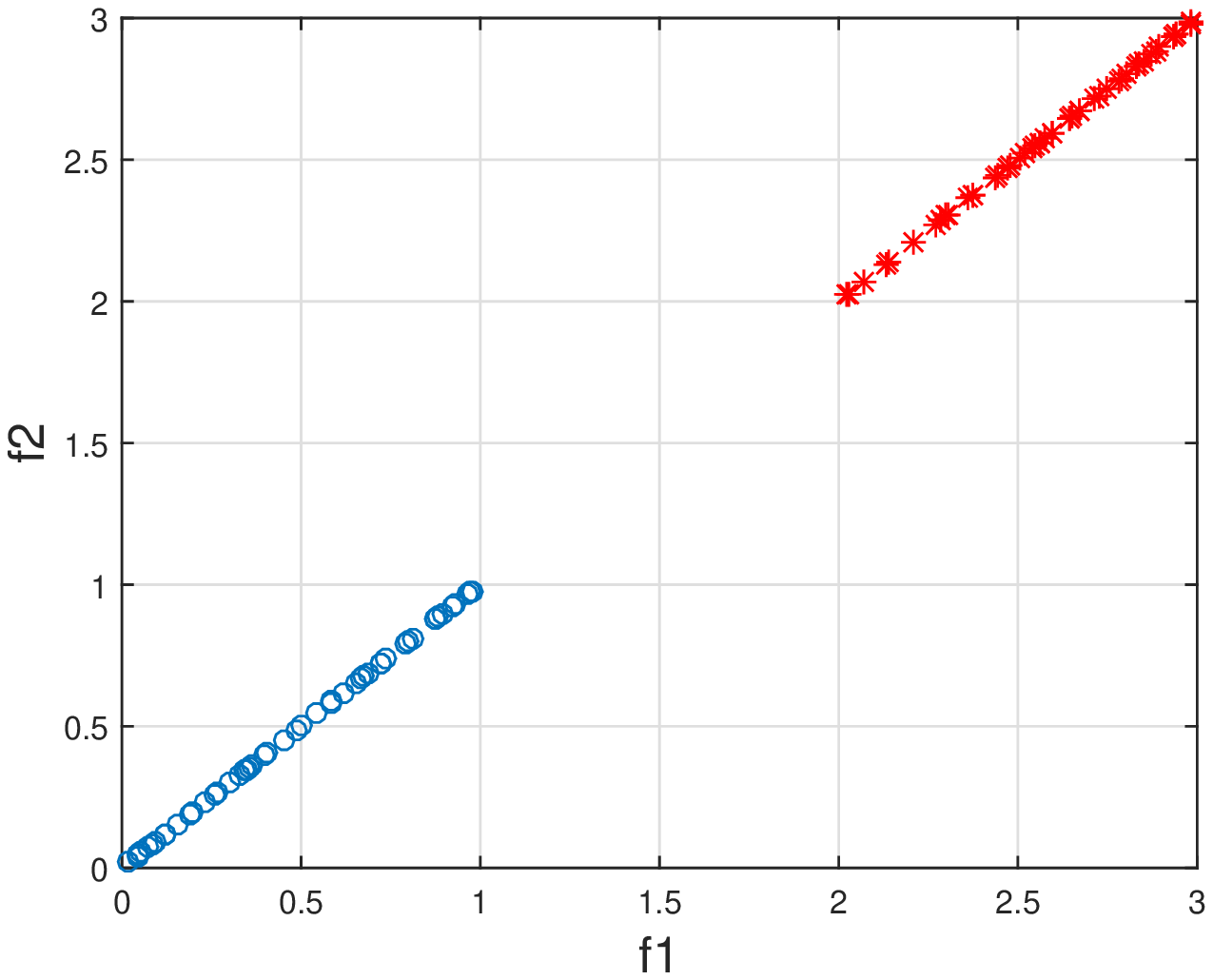}}
\end{minipage}
\begin{minipage}{0.4\textwidth}
\centering
\subfigure[irrelevant feature $f_{3}$\label{fig:featureIllustration-c}]
{\includegraphics[width=\textwidth]{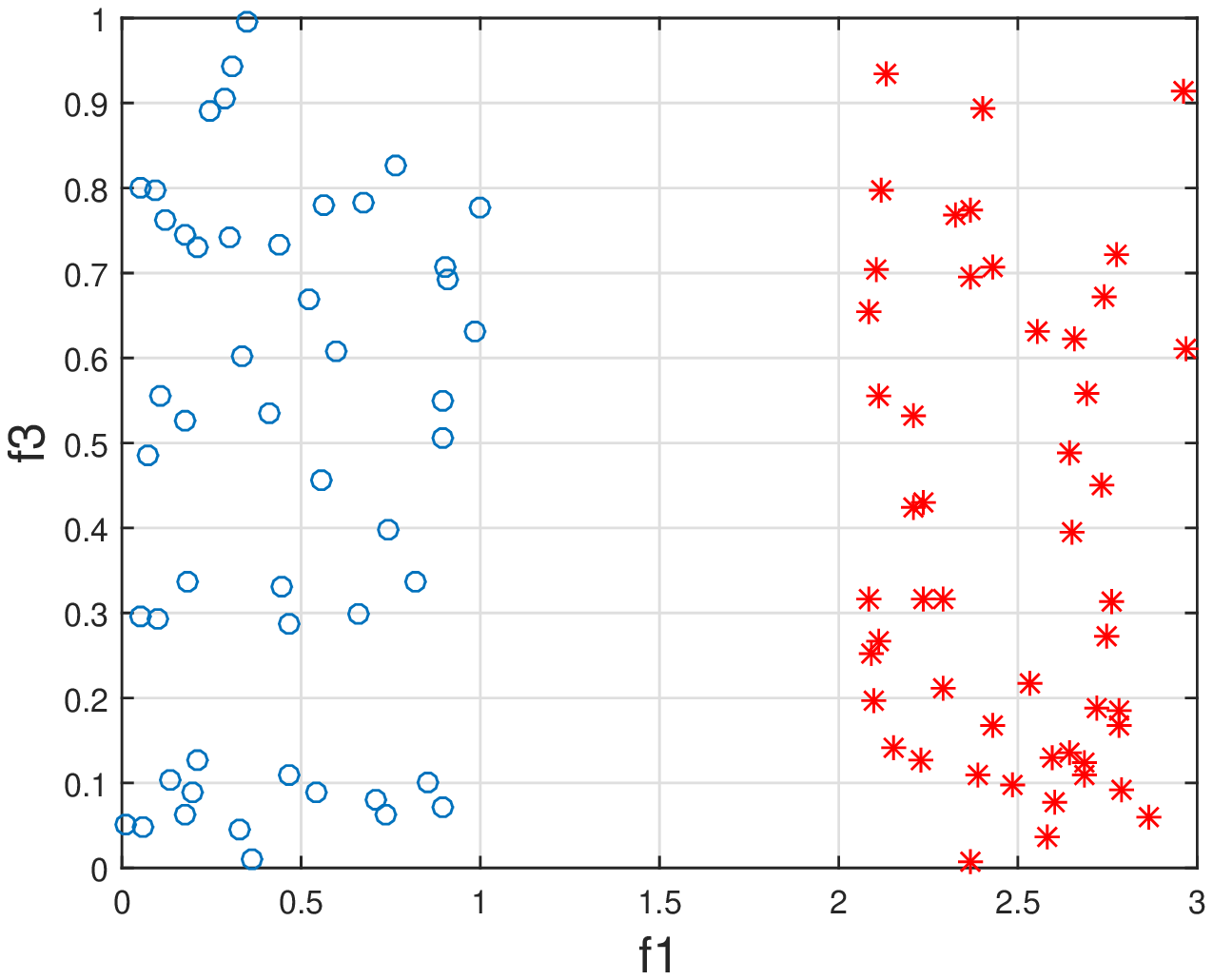}}
\end{minipage}
\centering
\caption{Example of relevant, irrelevant and redundant features.}
\label{fig:featureIllustration}
\end{figure}

Feature selection, as a type of dimension reduction technique, has been proven to be effective and efficient in handling high dimensional data~\cite{liu2007computational,li2016feature}. It directly selects a subset of relevant features for the model construction. Since feature selection keeps a subset of original features, one of its major merit is that it well maintains the physical meanings of the original feature sets, and gives better model readability and interpretability. Due to this particular reason, it is more widely applied in many real world applications such as gene analysis and text mining. Feature selection obtains relevant features by removing irrelevant and redundant features. The removal of these irrelevant and redundant features reduces the computational and storage costs without significant loss of information or negative degradation of the learning performance. Taking Figure~\ref{fig:featureIllustration} as an example, feature $f_{1}$ is a relevant feature which can separate two classes (clusters) in Figure~\ref{fig:featureIllustration-a}; while in Figure~\ref{fig:featureIllustration-b}, feature $f_{2}$ is considered as a redundant feature w.r.t feature $f_{1}$ since feature $f_{1}$ already can discriminate two classes (clusters) well; in Figure~\ref{fig:featureIllustration-c}, feature $f_{3}$ is an irrelevant feature as it does not contain useful information to separate two classes (clusters).

According to the availability of class labels, we can categorize feature selection algorithms into supervised and unsupervised methods. Supervised feature selection is usually taken as a preprocessing step for the classification/regression task. It chooses features that can discriminate data instances from different classes or regression targets. Since the label information is known a priori, relevance of a feature is normally assessed by its correlation with class labels. On the other hand, unsupervised feature selection is generally applied for the clustering task. Without class labels to guide feature selection, it evaluates feature importance by some alternative criteria such as data similarity, local discriminative information and data reconstruction error.



With regard to search strategies, feature selection algorithms can be divided into wrapper methods, filter methods and embedded methods. Wrapper methods typically use the learning performance of a predefined model to evaluate the feature relevance. Specifically, it repeatedly chooses a subset of features and then evaluates the learning performance with these selected features until the highest learning performance is obtained. Since it scans through the whole search space, it is slow and seldom used in practice. Filter methods, on the other hand, do not rely on any learning algorithms and are therefore more efficient. They exploit the characteristics of data to measure the feature relevance. Usually, they measure the scores of features based on some ranking criteria and then return the top ranked features. Since these methods do not explicitly consider the bias of learning algorithms, the selected features may not be optimal for a particular learning task. Embedded methods provide a trade-off solution between filter and wrapper methods which embed the feature
selection with the model learning, thus they inherit the merits of wrapper and filter methods: first, they include the interactions with the learning algorithm; and second, they are far more efficient than the wrapper methods since they do not need to evaluate feature sets iteratively.

\section{Challenges of Feature Selection}
Recently, the popularity of big data presents some challenges for the traditional feature selection task. Meanwhile, some unique characteristics of big data also bring about new opportunities for the feature selection research. In the next few subsections, we will present these challenges of feature selection for big data analytics from the following six aspects. In particular, the challenges of structured features, linked data, multi-source data and multi-view data, streaming data and features are from the perspective of data; while the last two challenges of scalability and stability, are from the performance perspective.

\subsection{Structured Features}
Most of existing feature selection algorithms are designed for generic data and they are based on a strong assumption that features do not have explicit correlations. In other words, they completely ignore the intrinsic structures among features. For example, these feature selection methods may select the same subset
of features even though the features are reshuffled~\cite{ye2012sparse}. In many real world applications, features exhibit various kinds of structures, e.g., spatial or temporal smoothness, disjoint groups, overlap groups, trees and graphs. When applying feature selection algorithms on the datasets with structured features, it is beneficial to explicitly incorporate this prior knowledge, which may improve post learning tasks such as classification and clustering. Next, we will focus on the most three common feature structures, i.e., group structure, tree structure and graph structure.

\begin{figure}[!t]
  \centering
    \includegraphics[width=0.6\textwidth]{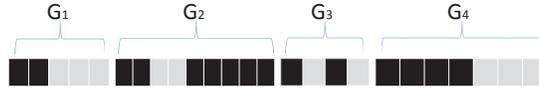}
      \caption{Group structures among features.}
\label{fig:group}
\end{figure}

The first structure features may exhibit is group structure. Examples of group structured features include different frequency bands represented as groups in signal processing~\cite{mcauley2005subband} and genes with similar functionalities acting as groups in bioinformatics~\cite{ma2007supervised}. Therefore, when performing feature selection, it is more appealing to explicitly take into consideration the group structure among features. Figure~\ref{fig:group} shows an illustrative example of features with group structures (4 groups).

\begin{figure}[!t]
  \centering
    \includegraphics[width=0.6\textwidth]{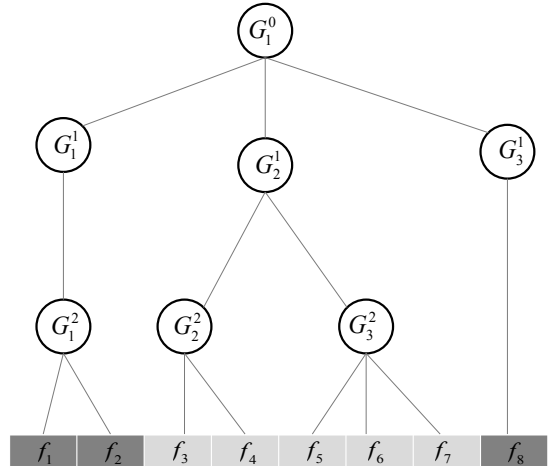}
      \caption{Tree structures among features.}
\label{fig:tree}
\end{figure}

In addition to the group structures, features can also form other kinds of structures such as tree structure. For example, in image processing such as face images, different pixels (features) can be represented as a tree, where the root node indicates the whole face, its child nodes can be the different organs, and each specific pixel is considered as a leaf node. In other words, these pixels enjoy a spatial locality structure. Figure~\ref{fig:tree} shows an example of 8 features with four layers of tree structure. Another motivating example is that genes/proteins may form certain hierarchical tree structures~\cite{jenatton2011structured}.
\begin{figure}[!t]
  \centering
    \includegraphics[width=0.6\textwidth]{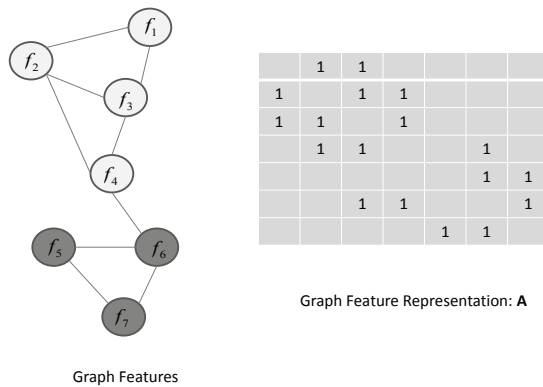}
      \caption{Graph structures among features.}
\label{fig:graph}
\end{figure}

Features may also form graph structures. For example, in natural language processing, if we take each word as a feature, we have synonyms and antonyms relationships between different words~\cite{fellbaum1998wordnet}. Moreover, many biological studies show that genes tend to work in groups according to their biological functions, and there are strong dependencies between some genes. Since features show some dependencies, we can model the features by an undirected graph, where nodes represent features and edges among nodes show the pairwise dependencies between features. An illustrative example of 7 features with graph structure is shown in Figure~\ref{fig:graph}.

\subsection{Linked Data}
\begin{figure}[!t]
\centering
\begin{minipage}{0.4\textwidth}
\centering
\subfigure[Linked data\label{fig:linkedfeature-a}]
{\includegraphics[width=\textwidth]{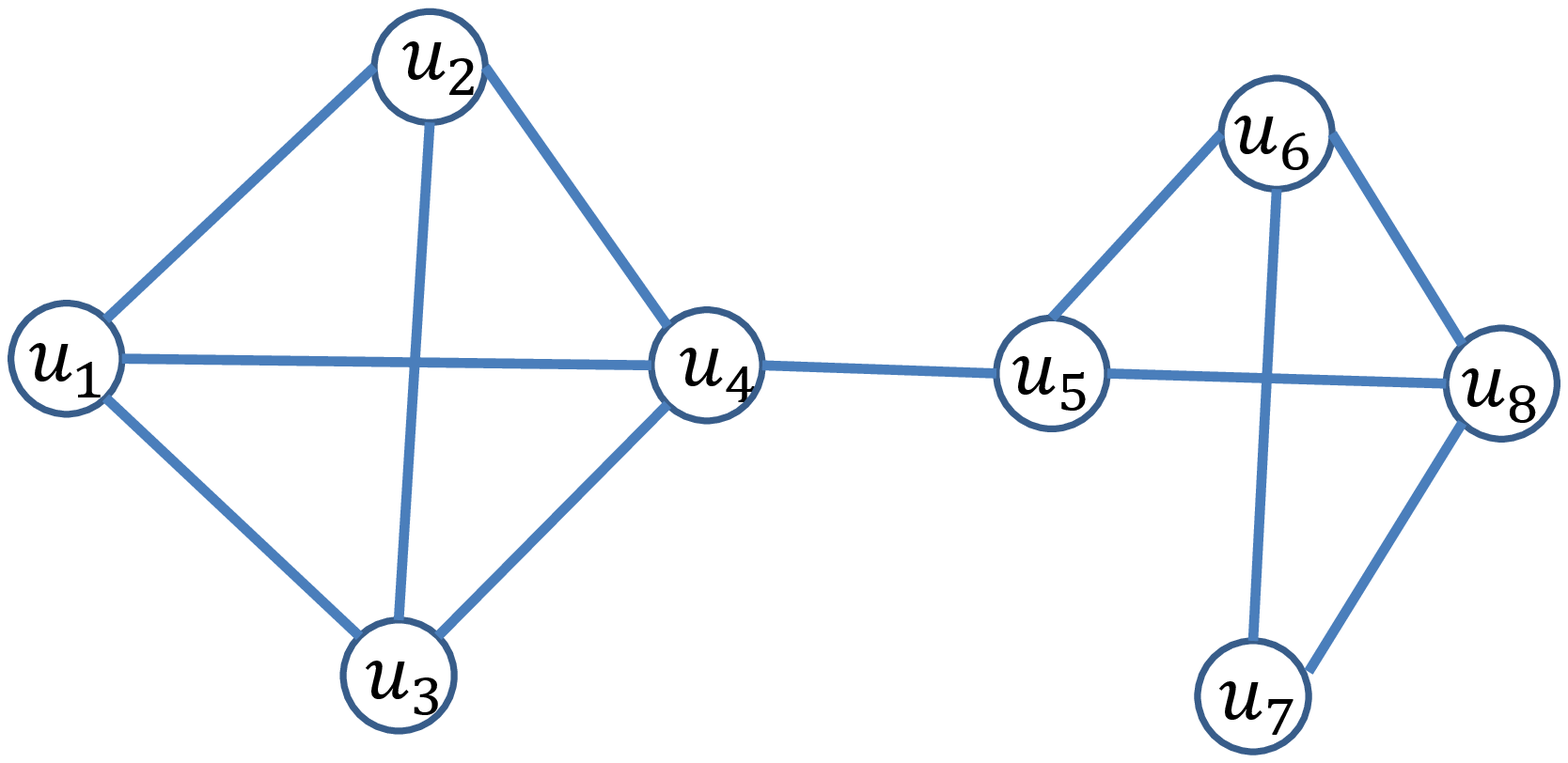}}
\end{minipage}
\begin{minipage}{0.4\textwidth}
\centering
\subfigure[Attribute-value data representation\label{fig:linkedfeature-b}]
{\includegraphics[width=\textwidth]{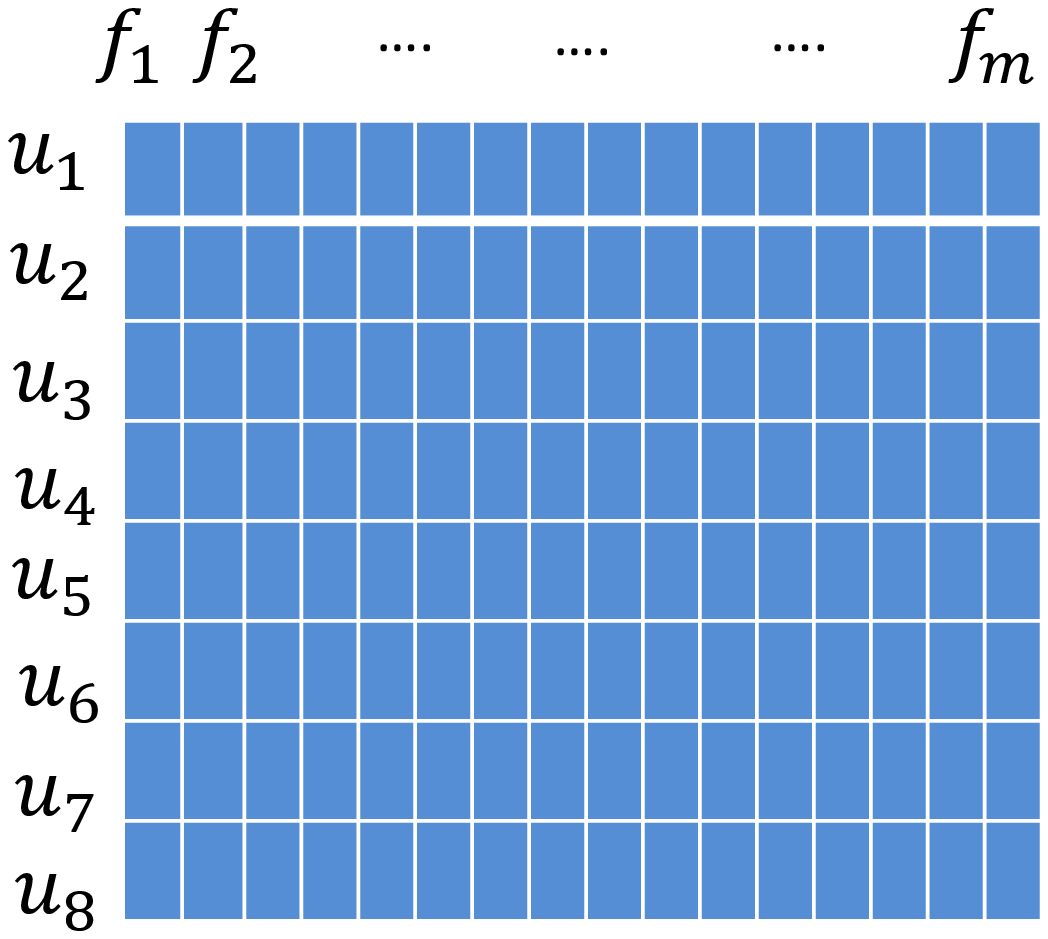}}
\end{minipage}
\begin{minipage}{0.8\textwidth}
\centering
\subfigure[Linked data representation\label{fig:linkedfeature-c}]
{\includegraphics[width=\textwidth]{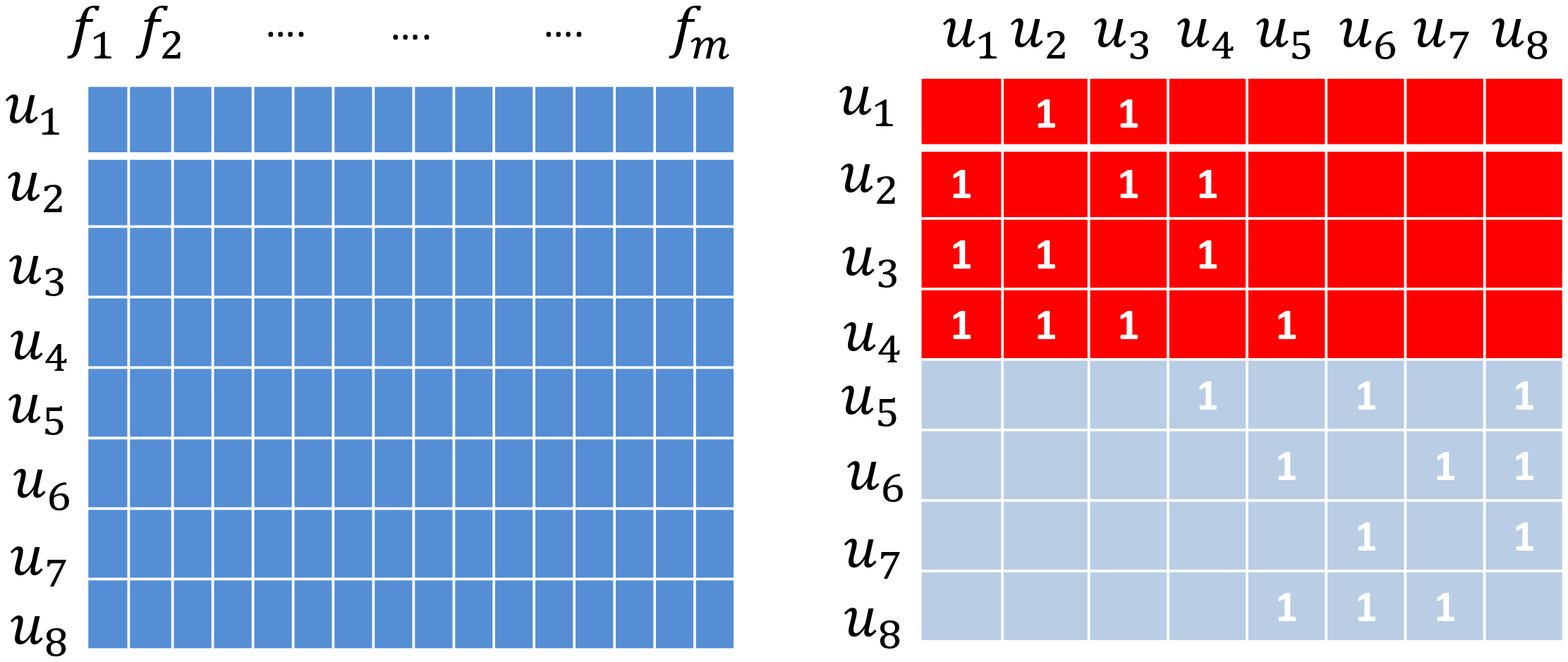}}
\end{minipage}
\centering
\caption{An illustrative example of linked data.}
\label{fig:linkedfeature}
\end{figure}

Linked data becomes ubiquitous in many platforms such as Twitter\footnote{https://twitter.com/} (tweets linked through hyperlinks), social networks in Facebook\footnote{https://www.facebook.com/} (users connected by friendships) and biological networks (protein interactions). Since linked data are correlated with each other by different types of links, they are distinct from traditional attribute-value data. Figure~\ref{fig:linkedfeature} presents an illustrative example of linked data and its two representations. Figure~\ref{fig:linkedfeature-a} shows 8 linked instances ($u_{1}$ to $u_{8}$) while Figure~\ref{fig:linkedfeature-b}
is a conventional representation of attribute-value data such that each row corresponds to one instance and each column corresponds to one feature. As mentioned above, linked data provides an extra source of information in the form of links, which can be represented by an adjacency matrix, illustrated in Figure~\ref{fig:linkedfeature-c}. The challenges of feature selection for linked data~\cite{li2016toward,li2016robust,tang2012unsupervised} lie in the following three aspects: (1) how to exploit relations among data instances; (2) how to take advantage of these relations for feature selection; and (3) linked data are often unlabeled, how to evaluate the relevance of features without the guide of label information.

\subsection{Multi-Source Data and Multi-View Data}
In many data mining and machine learning tasks, we have multiple data sources for the same set of data instances. For example, recent
advancement in bioinformatics reveal the existence of some non-coding RNA species in addition to the widely used messenger RNA, these non-coding RNA species functions across a variety of biological processes. The availability of multiple data sources makes it possible to address some problems otherwise unsolvable using a single source, since the multi-faceted representations of data can help depict some intrinsic patterns hidden in a single source of data. For multi-source feature selection, we usually have a target source and other sources complement the selection of features on the target source~\cite{zhao2011spectral}.

\begin{figure}[!t]
\centering
\begin{minipage}{0.7\textwidth}
\centering
\subfigure[Multi-source feature selection\label{fig:Multi-source}]
{\includegraphics[width=\textwidth]{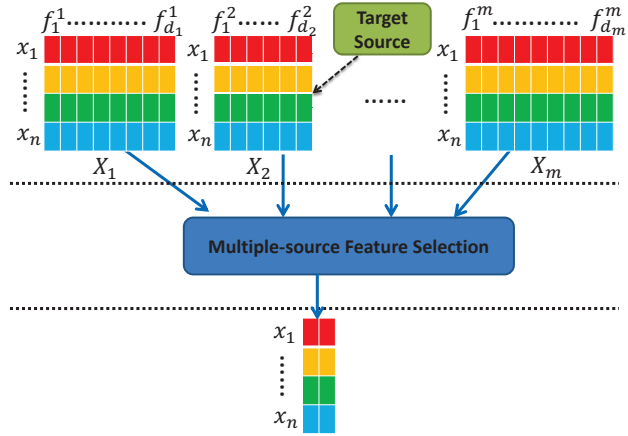}}
\end{minipage}
\begin{minipage}{0.7\textwidth}
\centering
\subfigure[Multi-view feature selection\label{fig:Multi-view}]
{\includegraphics[width=\textwidth]{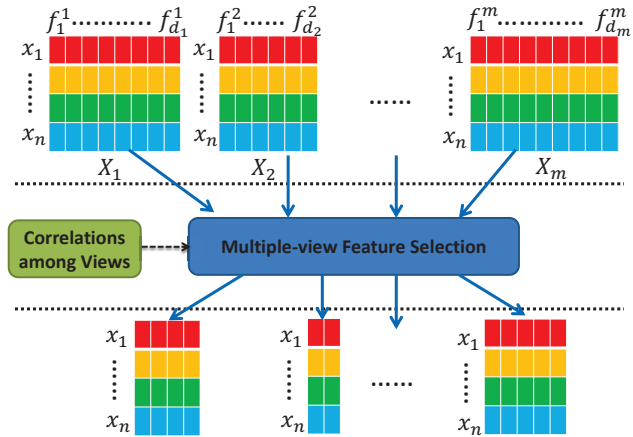}}
\end{minipage}
\caption{Differences between multi-source and multi-view feature selection.}
\label{fig:multi-source-multi-view}
\end{figure}

Multi-view sources represent different facets of data instances via different feature spaces. These feature spaces are naturally dependent and also high dimensional, which suggests that feature selection is necessary to prepare these sources for effective data mining tasks such as multi-view clustering. A task of multi-view feature selection thus arises, which aims to select features from different feature spaces simultaneously by using their relations~\cite{tang2013unsupervised,wang2013multi}. For example, we would like to select pixel features, tag features, and text features about images in Flickr\footnote{https://www.flickr.com/} simultaneously. Since multi-view feature selection is designed to select features across multiple views by using their relations, they are naturally different from multi-source feature selection. We illustrate the difference between multi-source feature selection and multi-view feature selection in Figure~\ref{fig:multi-source-multi-view}.

\subsection{Streaming Data and Features}
\begin{figure}[!htbp]
\centering
\includegraphics[width=0.5\textwidth]{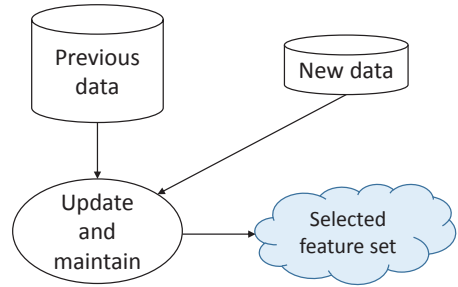}
\caption{A framework of feature selection on data streams.}
\label{fig:datastreams}
\end{figure}

In many scenarios, we are faced with a significant amount of data which need to be processed in a real time to gain insights. In the worst cases, the size of data or the features are unknown or even infinite, thus it is not practical to wait until all data instances or features are available to perform feature selection. For streaming data, one motivating example is that in online spam email detection problem, new emails are constantly arriving, it is not easy to employ batch-mode feature selection methods to select relevant feature in a timely manner. Therefore, some feature selection algorithms are proposed to maintain and update a feature subset when new data streams are constantly arriving. The process of feature selection on data streams is illustrated in Figure~\ref{fig:datastreams}. In some settings when the streaming data cannot be loaded into the memory, one pass of the data is required. We can only make one pass of the data where the second pass is either unavailable or computational expensive. How to select relevant features timely by one pass of data~\cite{huang2015unsupervised} is still a challenging problem.

\begin{figure}[!htbp]
\centering
\includegraphics[width=0.7\textwidth]{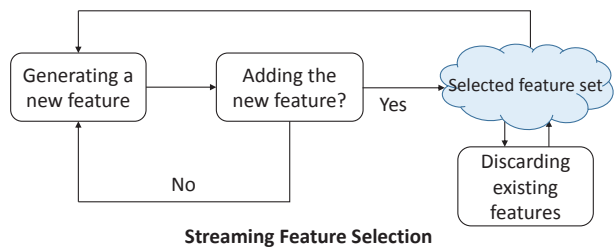}
\caption{A framework of streaming feature selection.}
\label{fig:StreamingFS}
\end{figure}

On an orthogonal setting, feature selection for streaming features also has its practical significance. For example, Twitter produces more than 320 millions of tweets everyday and a large amount of slang words (features) are continuously being generated. These slang words promptly grab users’ attention and become popular in a short time. Therefore, it is more preferable to perform streaming feature selection to rapidly adapt to the changes~\cite{li2015unsupervised}. A general framework of streaming feature selection is presented in Figure~\ref{fig:StreamingFS}. At each time step, a typical streaming feature selection algorithm first determines whether to accept the most recently arrived feature; if the feature is added to the selected feature set, it then determines whether to discard some existing features from the selected feature set. The process repeats until no new features show up anymore.

\subsection{Scalability}
With the tremendous growth of dataset sizes, the scalability of most current feature selection algorithms may be jeopardized. In many scientific and business applications, data are usually measured in terabyte (1TB = $10^{12}$ bytes). Normally, datasets in the scale of terabytes cannot be loaded into the memory directly and therefore limits the usability of most feature selection algorithms. Currently, there are some attempts to use distributed programming frameworks such as MapReduce and MPI to perform parallel feature selection for very large-scale datasets~\cite{singh2009parallel}. Recently, big data of ultrahigh dimensionality has emerged in many real-world applications such as text mining and information retrieval. Most feature selection algorithms does not scale well on the ultrahigh dimensional data, its efficiency deteriorates quickly or is even computational infeasible. In this case, well-designed feature selection algorithms in linear or sub-linear running time are more preferred.

\subsection{Stability}
The stability of these algorithms is also an important consideration when developing new feature selection algorithms~\cite{he2010stable}. A motivating example from the field of bioinformatics shows that domain experts would like to see the same set or similar set of genes (features) to be selected each time when they obtain new samples in the small amount of perturbation. Otherwise domain experts would not trust these algorithms when they get quite different sets of features with small data perturbation. It is also found that the underlying characteristics of data may greatly affect the stability of feature selection algorithms and the stability issue may also be data dependent. These factors include the dimensionality of feature, number of data instances, etc. In against with supervised feature selection, stability of unsupervised feature selection algorithms has not be well studied yet. Studying stability for unsupervised feature selection is much more difficult than that of the supervised methods. The reason is that in unsupervised feature selection, we do not have enough prior knowledge about the cluster structure of the data. Thus we are uncertain that if the new data instance, i.e., the perturbation belongs to any existing clusters or will introduce new clusters.

\section{Feature Selection Repository}
To tackle the challenges of feature selection for big data analytics and to promote the feature selection research in this community, we present an open-source feature selection repository - \emph{scikit-feature} (http://featureselection.asu.edu/). The purpose of this feature selection repository is to collect some widely used feature selection algorithms that have been developed in the feature selection research to serve as a platform for facilitating their application, comparison and joint study. The feature selection repository also effectively assists researchers to achieve more reliable evaluation in the process of developing new feature selection algorithms.

Currently, \emph{scikit-feature} consists of popular feature selection algorithms in the following categories:
\begin{itemize}
\setlength\itemsep{0.3em}
\item Similarity based feature selection
\item Information theoretical based feature selection
\item Statistical based feature selection
\item Sparse learning based feature selection
\item Wrapper based feature selection
\item Structural feature selection
\item Streaming feature selection
\end{itemize}

Among these different categories of feature selection methods, similarity based, information theoretical based, and statistical based methods correspond to the filter methods discussed above. Wrapper based methods and sparse learning based methods correspond to the wrapper methods and embedded methods, respectively. We also include structural features, linked data, multi-view and multi-source data to the category of structural feature selection, and streaming data and features to the streaming feature selection category.

In addition, scikit-feature also provides many benchmark feature selection datasets, and evaluation examples on how to evaluate feature selection algorithms via classification or clustering task. The experimental results can be obtained from our repository project website (http://featureselection.asu.edu/datasets.php).
For each dataset, we list all applicable feature selection algorithms along with its evaluation on either classification or clustering task. We also provide an interactive tool FeatureMiner~\cite{cheng2016featureminer} to ease the usage of these feature selection algorithms based on the repository.

\section*{Acknowledgements}
This material is, in part, supported by National Science Foundation (NSF) under grant number IIS-1217466.

\end{document}